# CEScore: Simple and Efficient Confidence Estimation Model for Evaluating Split and Rephrase

**AlMotasem Bellah Al Ajlouni**
School of Computer Science and Technology
University of Science and Technology of China
Hefei, China
motasem@mail.ustc.edu.cn

**Jinlong Li**
School of Computer Science and Technology
University of Science and Technology of China
Hefei, China
jlli@ustc.edu.cn

## Abstract

The split and rephrase (SR) task aims to divide a long, complex sentence into a set of shorter, simpler sentences that convey the same meaning. This challenging problem in NLP has gained increased attention recently because of its benefits as a pre-processing step in other NLP tasks. Evaluating quality of SR is challenging, as there no automatic metric fit to evaluate this task. In this work, we introduce CEScore, as novel statistical model to automatically evaluate SR task. By mimicking the way humans evaluate SR, CEScore provides 4 metrics ($S_{score}$, $G_{score}$, $M_{score}$, and $CE_{score}$) to assess simplicity, grammaticality, meaning preservation, and overall quality, respectively. In experiments with 26 models, CEScore correlates strongly with human evaluations, achieving 0.98 in Spearman correlations at model-level. This underscores the potential of CEScore as a simple and effective metric for assessing the overall quality of SR models.

## 1 Introduction

Splitting the long sentence into many shorter and simpler sentences was one of the text simplification (TS) techniques until 2017, when Narayan et al. classified it as a distinct NLP task named Split and Rephrase (SR). Since that, more attention has emerged to it as a new challenging problem in NLP.

According to previous research (Narayan et al., 2017; Stajner et al., 2016; Saha, 2018; Miwa et al., 2010), many NLP tasks benefit from SR as a pre-processing step to improve performance, especially the tasks whose expected quality is negatively correlated with sentence length, such as machine translation, information extraction, and text parsing. However, improving SR models hinges on our capability to accurately and efficiently evaluate the quality of their outputs.

The absence of dedicated automatic evaluation metrics for SR has resulted in the adaptation of metrics originally designed for assessing Machine Translation (MT). This approach arises from the recognition that SR essentially involves monolingual translation, where a complex sentence is transformed into simpler sentences within the same language. Prominent evaluation metrics such as BLEU (Papineni and Roukos, 2002) and BERT score (Zhang et al., 2019), have been repurposed for this task. However, based on previous researches (Xu et al., 2016; Sulem et al., 2018 a; Sulem et al., 2018 b), the use of these standards has not proven entirely effective in evaluating SR, even when good references are available. This inadequacy stems from a fundamental issue: these metrics tend to favor longer sentences, which directly contradicts the essence of SR, where the goal is to create shorter sentences. Consequently, this led the researcher to relay on manual evaluation that utilized humans to properly evaluate the output of a SR models. This represents a big barrier of improving these systems, as relying solely on human evaluation can be time-consuming. Furthermore, the results of human evaluation are subjective and can vary depending on the evaluators' backgrounds and biases.

This paper presents a novel statistical model called CEScore that functions as a ***Confidence Estimation Score*** model, It directly evaluates the quality of SR by considering three fundamental dimensions: Simplicity (S), ensuring that the text becomes more straightforward; Meaning preservation (M), verifying that the essence of the original content remains intact; and Grammaticality (G), assessing the text's adherence to proper grammar.

CEScore generates four distinct scores: $S_{score}$, $M_{score}$, $G_{score}$, and $CE_{score}$, each of which represents the model's assessment for a specific criterion. This approach mirrors the way humans naturally evaluate SR, providing a more contextually relevant and interpretable assessment of the quality of SR.

In our quest to compute $S_{score}$, we have developed new statistical formulas for evaluating sentence simplicity. These formulas, known as the Sentence Length Score (SLS), Average Sentence Familiarity (ASF), and Text Simplicity Score (TSS), designed to evaluate text simplicity based on many factors influence the simplicity of a text, such as word count, clause structure, word count within each clause, and the familiarity of vocabulary used.

To test CEScore's effectiveness, we compared its metrics with others metrics that commonly used for evaluating SR systems, including BLEU , SARI (Xu et al., 2016) , BERTscore, and SAMSA (Sulem et al., 2018 b). Our comparison utilizes a sizable human evaluation benchmark provided by Sulem et al. (2018 c) as a foundational reference for this comparison. We performed that comparison of two levels: sentence-level and model-level.

Our research contributes significantly to the field of SR evaluation in the following key aspects:

1. **Statistical Functions for Evaluation:** We introduce a set of statistical functions and algorithms that facilitate the evaluation of SR. These functions, namely SScore, GScore, and MScore, provide a comprehensive assessment of SR in terms of S, G, and M, respectively.
2. **Innovative Formulas for Simplicity Assessment:** In our pursuit of evaluating the simplicity aspect, we have developed novel statistical formulas. These include the Sentence Length Score (SLS), Average Sentence Familiarity (ASF), and Text Simplicity Score (TSS), which offer valuable insights into the simplicity of SR models' output.
3. **CEScore:** We create and make accessible[1][2] a non-referenced automatic evaluation metric, CEScore, designed to assess SR in a manner akin to human evaluation. CEScore evaluates SR based on the critical criteria of S, M, and G. CEScore, demonstrates a remarkable level of agreement with human judgments, particularly at the model level. It achieves impressive coefficient values of 0.98 and 0.95 in Spearman and Pearson correlations, respectively, underscoring CEScore's potential as a reliable and effective evaluation metric for SR models.

## 2 Related Work

SR models are evaluated using a combination of automatic and manual evaluation methods to assess their effectiveness in making text more understandable and easier to read.

### 2.1 Manual evaluation Method

The absence of a robust automated evaluation metric for SR has led researchers to rely on manual evaluation methods (Alva-Manchego et al., 2020), where human assessors carefully scrutinize the outcomes generated by SR models. Typically, the human evaluators assess the model's outputs based on three fundamental criteria (Narayan et al., 2017; Sulem et al., 2018 c; Niklaus et al., 2019):

**Grammaticality (G):** This criterion evaluates the degree of grammatical accuracy exhibited by the output in comparison to the input.

**Meaning Preservation (M):** This dimension estimates how effectively the output maintains the core meaning of the input.

**Simplicity (S):** This factor gauges the extent to which the output simplifies the input.

Annotators assign scores to each criterion individually, G and M are assessed on a scale of 1 to 5, while S is evaluated on a scale of -2 to 2, wherein zero signifies an equivalent degree of simplicity, a positive value indicates a simplified output relative to the input, and a negative value implies the converse (Narayan and Gardent, 2014; Nisioi et al., 2017).

While manual evaluation remains the gold standard for evaluating SR models, it is not without its limitations. This approach is inherently time-consuming, often spanning several months for completion, and it requires significant human labor resources that are not reusable (Papineni and Roukos, 2002). Furthermore, the inherent subjectivity in manual evaluation introduces variability into the results, stemming from the diverse backgrounds and biases of the evaluators (Xu et al., 2016). Consequently, when comparing

---

[1] https://github.com/motasemajlouni/CEScore

[2] Also, you can install the CEScore model by using ***"pip install CEScore"***

SR models, it becomes essential to re-evaluate model outputs using the same evaluators to mitigate the influence of evaluator biases.

## 2.2 Automatic Evaluation Method

In the realm of automatic evaluation metrics, two distinct approaches have emerged (Sulem et al., 2018 c). The first approach relies on references, wherein the output of a TS model is evaluated by comparing it to reference texts or human-generated simplifications. Metrics like BLEU, SARI, and BERTscore fall under this category.

BLEU is an n-gram-based evaluation metric widely used for assessing MT quality. It has been employed as an automatic metric for evaluating SR models (Narayan et al., 2017; Aharoni et al., 2018; Botha et al., 2018; Niklaus et al., 2019). While BLEU exhibits a strong correlation with human judgments regarding G and M, studies have shown its limitations in predicting S, whether in terms of lexical simplification (Xu et al., 2016) or structural simplification (Sulem et al., 2018 a).

In response to BLEU's limitations, Xu et al. (2016) introduced SARI as a reference metric specifically designed for the evaluation of text simplification (TS) systems that emphasize lexical simplification. It compares the n-grams in the system output with those in the input and human references. However, Sulem et al. (2018 a) found that neither BLEU nor SARI are well-suited for assessing models that involve structural simplification. Consequences, they introduced SAMSA as metric primarily tailored for the evaluation of structural simplification.

BERTScore introduced by Zhang et al., in 2019, for evaluating the quality of machine-generated text. It is specifically designed to measure the similarity between the generated text and a reference text using contextual embeddings from BERT (Bidirectional Encoder Representations from Transformers), a popular pre-trained NLP model (Kenton et al., 2019). Recently, BERTScore has gained popularity as an evaluation metric for TS tasks, as it is well-suited for assessing models that involve both of structural and lexical simplification, as reported by AlAjlouni et al. (2023 ).

The second approach, on the other hand, does not necessitate the presence of reference texts. These metrics are often referred to as Confidence Estimation (CE) or Quality Estimation (QE) (Blatz et al., 2004). They evaluate models without the need for reference comparisons. This approach is particularly useful when references are scarce or unavailable, providing a more versatile and practical means of assessing the performance of models.

This concept initially emerged within the domain of MT for evaluating the quality of automatically translated text (Blatz et al., 2004; Specia et al., 2009; Martins et al., 2017; Specia et al., 2018 ). In this field, early research by Blatz et al. (2004) treated CE as a binary classification problem, aiming to distinguish between "good" and "bad" translations .Conversely, Specia et al., (2009) approached CE as a regression task, wherein they aimed to estimate a continuous quality score for each translated sentence .

In the realm of TS, Sulem et al. in 2018 introduced SAMSA as first reference-less automatic metric to address structural aspects of TS. It employs semantic parsing to assess the quality of simplification by breaking down the input text based on its scenes and comparing it to the output. SAMSA penalizes cases where the number of sentences in the output is more than the number of scenes in the input. $SAMSA_{abl}$ is a modified version of SAMSA in which the penalization condition is omitted. However , SAMSA relies on the TUPA parser (Hershcovich et al., 2017) to decompose source sentences into their constituent scenes. This dependency has limited the practicality of SAMSA, primarily due to the parser's slow processing speed and the intricate preparation steps involved. Furthermore, the parser's accuracy is often compromised, given the complexity of the language, which has adversely affected the accuracy of SAMSA. As a solution, the authors resorted to manually decomposing source sentences, but this adaptation eliminates SAMSA advantageous as a reference-less automatic metric.

CEScore is a reference-less automated metric that emulates the manual evaluation process. It assesses the SR process based on criteria such as simplicity, grammaticality, and meaning preservation. CEScore calculations are simple and fast, as they involve statistical analysis of linguistic components in both the source sentence and the simplified sentence.

## 3 CEScore Model

The CEScore model, where CEScore stands for *Confidence Estimation Score*, represents a statistical model tailored to assess the quality of SR without requiring reference texts for comparison. This model emulates the human approach. It generates four distinct scores: $S_{score}$, $M_{score}$, $G_{score}$, and $CE_{score}$, each of which represents the model's assessment for specific criteria within the simplification process.

The $S_{score}$ indicates the degree of simplicity achieved in the simplified text when contrasted with the complexity of the original text. The $M_{score}$ quantifies the extent to which the meaning of the original complex text is preserved. The $G_{score}$ assesses the grammatical correctness of the simplification. Lastly, the $CE_{score}$ serves as an overarching quality estimation, offering a holistic evaluation of the SR process. Table 1 presents CEScore model steps, and here's a step-by-step description of how it works:

The CEScore Model takes two input texts: the complex text ($\mathcal{T}_C$) and the simplified text ($\mathcal{T}_S$) *(line 1)*. It first computes the $M_{score}$ by calling the **MScore** function, which evaluates how well the simplified text preserves the meaning of the complex text *(line 2)*. Next, it calculates the $G_{score}$ by calling the **GScore** Function, which assesses the grammatical correctness of the simplified text *(line 3)*. Similarly, it computes the $S_{score}$ by calling the SScore function, which measures how effectively the simplified text achieves simplicity *(line 4)*.

The $CE_{score}$ is calculated by taking the geometric mean of the three scores: $S_{score}$, $M_{score}$, and $G_{score}$. This provides an overall quality estimation for the TS, considering all three criteria (S, M, and G) *(line 4)*. Finally, the CEScore Model returns four scores: $S_{score}$, $M_{score}$, $G_{score}$ and $CE_{score}$. These scores are all represented as values between 0 and 1, where a score above 0.5 is considered acceptable, and a score below 0.5 indicates a failure to meet the criteria.

### 3.1 SScore Function

The SScore is a statistical function designed to estimate S criterion, which plays a crucial role in our CEScore model, considering that the main goal of the SR is to produce simplified sentences. Evaluating S is a challenging task, as there are many factors influence the simplicity of a text, such as word count, clause structure, word count within each clause, and the familiarity of vocabulary used. However, as high-lighted by Martin et al. (2018), sentence length is a highly effective metric that exhibits a strong correlation with human evaluations for the S criterion.

To the best of our knowledge, there is currently no effective metric available for estimating the simplicity of sentences. As a result, human evaluation remains the sole reliable method for assessing sentence simplicity. In our quest to estimate simplicity effectively, we have developed new statistical formulas for evaluating sentence simplicity. These formulas, known as the Sentence Length Score (SLS), Average Sentence Familiarity (ASF), and Text Simplicity Score (TSS).

The **Sentence Length Score (SLS)** is a statistical formula developed to encapsulate the connection we uncovered between sentence length and the S scores as assessed in manual evaluation. Our analysis revealed intriguing pattern. What we observed was that once a sentence dropped below a specific lower word count threshold, it consistently appeared as simple in the eyes of human evaluators, regardless of its actual length. Conversely, when a sentence exceeded an upper word count threshold, it consistently appeared complex, even if its length was not particularly long. For sentences that fell within the range between these two thresholds, we discerned a linear relationship between their length and the perceived level of simplicity.

It's important to acknowledge that these thresholds can vary among individuals based on their linguistic proficiency. However, based on our empirical findings, we determined that the most suitable values for these thresholds are 5 words for the lower threshold and 25 words for the upper threshold. SLS, as described in Equation 1, is crafted to capture and mirror this observed pattern effectively.

| Algorithm 1: CEScore Model |
|---|
| 1: **Function** $CEScore(\mathcal{T}_C, \mathcal{T}_S)$ |
| 2: $\quad \mathcal{M}_{score} \leftarrow MScore(\mathcal{T}_C, \mathcal{T}_S)$ |
| 3: $\quad G_{score} \leftarrow GScore(\mathcal{T}_C, \mathcal{T}_S)$ |
| 4: $\quad S_{score} \leftarrow SScore(\mathcal{T}_C, \mathcal{T}_S)$ |
| 5: $\quad CE_{score} = \sqrt[3]{G_{score} \times \mathcal{M}_{score} \times S_{score}}$ |
| 6: **Return** $\mathcal{M}_{score}, S_{score}, G_{score}, CE_{score}$ |

Table 1: shows the steps that the CEScore model performs to calculate the $S_{score}$, $M_{score}$, $G_{score}$ and $CE_{score}$.

$$SLS(S) = 1 - \left(\frac{1}{1+e^{(-\tau(|S_{tokens}|-\omega))}}\right) \quad (1)$$

In this equation, $S_{tokens}$ is a list of tokens (words) that are belong to the sentence $S$ after removing non-alphabetic and non-numerical tokens. The constants $\tau$ and $\omega$ are used to control the threshold values, which are set to 0.22 and 13 for $\tau$ and $\omega$, respectively.

The **Average Sentence Familiarity (ASF)** is a novel formula we have developed to measure the familiarity of a sentence for a broad audience. To accomplish this, we have incorporated two key scales: the 'percent_known' scale from Brysbaert's concreteness list (Brysbaert et al., 2014) and the 'Zipf-value' scale from the SUBTLEX-US frequency list (Heuven et al., 2014).

In Brysbaert's 2014 concreteness list, the 'percent_known' value for a word signifies the percentage of participants who recognized that word. Conversely, the 'Zipf-value' scale is a frequency measure extracted from the SUBTLEX-US frequency list. This scale offers a more accessible way to comprehend word frequency compared to traditional measures. Zipf values range from 1 to 7, with values 1-3 denoting low-frequency words and values 4-7 indicating high-frequency words.

High-frequency words are those that appear frequently in everyday language and are likely to be encountered by readers in various contexts. These words are more likely to be stored in readers' mental lexicons, making them easier to process and recognize. We argue that frequent words are more likely to be familiar to readers, and can make a text more accessible and easier to understand.

Table 2 presents the algorithm that we follow to calculate the ASF value, which can be outlined as follows. Initially, it tokenizes the input sentence $S$ into a set of individual tokens and stores these tokens in the set $S_{tokens}$ using the Split function *(line 2)*. Subsequently, the algorithm defines another set, $\psi$, comprised of words sourced from the SUBTLEX-US frequency list, encompassing commonly recognized terms by a wide-ranging audience *(line 3)*. The intersection of $S_{tokens}$ and $\psi$ is computed and represented as $Š$, encapsulating tokens that are shared between the original sentence and the list of common words *(line 4)*. The audience familiarity score ($\mathcal{ASF}_{score}$) is then calculated as the arithmetic mean of three functions applied to each token in $Š$ *(line 5)*. These functions for each token $t$ in $Š$ are calculated as follows:

| Algorithm 2: ASF Formula | |
|---|---|
| 1: | **Function** $ASF(S)$ |
| 2: | $S_{tokens} = \{s \mid s \in Split(S)\}$ |
| 3: | $\psi = \{w \mid w \in SUBTLEX - US\}$ |
| 4: | $Š = S_{tokens} \cap \psi$ |
| 5: | $\mathcal{ASF}_{score} = \frac{1}{|Š|} \sum_{t \in Š} \frac{\hbar(t) \times z(t)}{s(t)}$ |
| 6: | **Return** $\mathcal{ASF}_{score}$ |

Table 2: shows the steps that the ASF formula performs to calculate audience familiarity score ($\mathcal{ASF}_{score}$), which measure the familiarity of a sentence $S$ for a broad audience.

| Algorithm 3: TSS Formula | |
|---|---|
| 1: | **Function** $TSS(T)$ |
| 2: | $\mathbb{S} = \{s \mid s \in ToSentences(T)\}$ |
| 3: | $\mathcal{F}_{lexl} = ASF(T) + 5\sqrt[3]{SLS(T)}$ |
| 4: | $\mathcal{F}_{strc} = \min_{s \in \mathbb{S}} ASF(s) \times SLS(s)$ |
| 5: | $\mathcal{TSS}_{Score} = \alpha \times \mathcal{F}_{lexl} + \beta \times \mathcal{F}_{strc}$ |
| 6: | **Return** $\mathcal{TSS}_{score}$ |

Table 3: shows the steps that the TSS formula performs to calculate $\mathcal{TSS}_{score}$, which represent the simplicity of a text $t$.

- $\hbar(t)$ represents the 'percent_known' value of the token $t$, which is based on Brysbaert's 2014 concreteness list.
- $z(t)$ represents the 'Zipf-value' value of the token $t$, which is based on the SUBTLEX-US frequency list.
- $s(t)$ represents the number of syllables of the token $t$.

The resulting $\mathcal{ASF}_{score}$ is then returned as the result *(line 6)*, reflecting the degree of audience familiarity with the input sentence.

The **Text Simplicity Score (TSS)** is an innovative formula devised to measure the simplicity of a given text. It combines two key elements: the ASF, which evaluates lexical simplicity, and the SLS, a pivotal factor in measuring simplicity according to human judgment.

The TSS is tailored to emphasize structural simplicity by considering the division of the text under evaluation into sentences, as outlined in Table 3. Here's an in-depth explanation of the TSS formula's operations:

The process commences by dividing the input text $T$ into individual sentences using the NLTK package (Loper and Bird, 2002), which are collected in the set $\mathbb{S}$ *(line 2)*. Subsequently, the algorithm calculates $\mathcal{F}_{lexl}$, signifying the estimated score for lexical simplicity *(line 3)*. This score is a composite of two components:

- **ASF(T):** This component represents the ASF for the entire text $T$, reflecting the degree of familiarity of the text's vocabulary with a broad audience.

- **5 times the cube root of SLS(T):** To accentuate the impact of SLS, we apply the cube root operation, while balancing its contribution to $\mathcal{F}_{lexl}$ by multiplying the cube root of SLS($T$) by a constant factor of 5.

Furthermore, the algorithm computes $\mathcal{F}_{strc}$, which signifies the estimated score for structural simplicity *(line 4)*. This score assesses the sentence considered the most complex among all sentences within the set $\mathbb{S}$. The determination is made by identifying the minimum value, calculated as the product of the ASF of each sentence $\mathbb{s}$ in $\mathbb{S}$ by its corresponding SLS. It is crucial to highlight that we use the minimum value instead of summation to reward texts consisting of multiple sentences.

The overall text's simplicity ($\mathcal{TSS}_{score}$) provides an estimation of the text's simplicity by considering both lexical and structural simplification aspects. It is computed as a linear combination of $\mathcal{F}_{lexl}$ and $\mathcal{F}_{strc}$, where α and β are adjustable coefficients to control the relative importance of lexical and structural simplicity in the overall simplicity score *(line 5)*. Although the weights α and β can be adjusted to customize the formula for specific simplicity evaluations, our empirical findings have led to the determination of optimal values for these weights, which are 0.45 and 0.55 for α and β, respectively.

It's noteworthy that the $\mathcal{TSS}_{score}$ typically falls within the range of 7 to 2.5 for the majority of texts. In theory, the extreme minimum value for $\mathcal{TSS}_{score}$, representing the most complex text, is approximately around 0.35. However, such a low value is exceedingly rare and would occur only in exceptional cases. To reach this minimum value, a text would need to consist of just one sentence with very lengthy words (more than 30 words), and none of these words would be found in the SUBTLEX-US frequency list.

On the other end of the spectrum, the extreme maximum value for $\mathcal{TSS}_{score}$ is 9.39, which is also a very rare occurrence. In this scenario, the text would be composed of a single token, *'the,'* which is the most widely recognized and frequently used word in the English language.

The SScore function employs the TSS formula to evaluate the simplicity of $\mathcal{T}_C$ and $\mathcal{T}_S$, as depicted in Table 4. The SScore commences by calculating the TSS score for the simplified text ($\mathcal{T}_S$) and the complex text ($\mathcal{T}_C$) *(lines 2 and 3)*. The algorithm then calculates a parameter, denoted as $\mathcal{D}$, which measures the disparity in simplicity between $\mathcal{T}_S$ and $\mathcal{T}_C$. $\mathcal{D}$ is determined as the relative difference between the TSS scores of the simplified and complex texts, normalized within the range [0, 1] *(line 4)*.

In the majority of cases, the value of $\mathcal{D}$ falls within the range [0, 1], indicating the typical relationship between $\mathcal{T}_S$ and $\mathcal{T}_C$. However, in extremely rare cases, $\mathcal{D}$ may deviate from this range. This occurs when the TSS score for $\mathcal{T}_S$ is at one extreme end of the TSS range, while the TSS score for $\mathcal{T}_C$ is at the opposite extreme. In these exceptional cases, the value of $\mathcal{D}$ may exceed the [0, 1] range. As a result, the determination of the $S_{score}$ (line 5) is based on the value of $\mathcal{D}$ as follows:

- If $\mathcal{D}$ falls within the range [0, 1], $S_{score}$ takes on the value of $\mathcal{D}$.

- If $\mathcal{D}$ is less than 0, $S_{score}$ is set to 0.

- If $\mathcal{D}$ exceeds 1, $S_{score}$ is set to 1.

Ultimately, the SScore function returns the $S_{score}$ *(line 6)*.

### 3.2 MScore Function

After conducting extensive testing to explore various methods for effectively assessing the M criterion, we have identified the most suitable approach. This approach involves counting the common words between the two texts, but we recognize that not all words are equal in their importance for preserving the text's meaning. The SR process often involves omitting less critical words, substituting complex terms with simpler alternatives, or rearranging and repeating nouns when splitting lengthy sentences into shorter ones.

| **Algorithm 4: SScore Function** | |
|---|---|
| 1: | **Function** $SScore\ (\mathcal{T}_C, \mathcal{T}_S)$ |
| 2: | $\mathcal{S}_{tss} \leftarrow TSS(\mathcal{T}_S)$ |
| 3: | $\mathcal{C}_{tss} \leftarrow TSS(\mathcal{T}_C)$ |
| 4: | $\mathcal{D} = \left(\dfrac{\mathcal{S}_{tss} - \mathcal{C}_{tss}}{\mathcal{S}_{tss} + \mathcal{C}_{tss}}\right) + 0.5$ |
| 5: | $S_{score} = \begin{cases} \mathcal{D}, & 0 \leq \mathcal{D} \leq 1 \\ 0, & 0 > \mathcal{D} \\ 1, & 1 < \mathcal{D} \end{cases}$ |
| 6: | **Return** $S_{score}$ |

Table 4: shows the steps that the SScore function performs to calculate $S_{score}$

| **Algorithm 5: MScore Function** | |
|---|---|
| 1: | **Function** $MScore\ (\mathcal{T}_C, \mathcal{T}_S)$ |
| 2: | $\mathcal{C} = \{c \mid c \in Split(\mathcal{T}_C)\}$ |
| 3: | $\mathcal{S} = \{c \mid c \in Split(\mathcal{T}_S)\}$ |
| 4: | $\mathcal{I} = \mathcal{C} \cap \mathcal{S}$ |
| 5: | $\mathcal{U} = \mathcal{C} \cup \mathcal{S}$ |
| 6: | $\mathcal{M}_{score} = \sum_{t \in \mathcal{I}} \dfrac{1}{1+z(t)} \Big/ \sum_{t \in \mathcal{U}} \dfrac{1}{1+z(t)}$ |
| 7: | **Return** $M_{score}$ |

Table 5: shows the steps that the MScore function performs to calculate $M_{score}$

To address these variations, we have devised a method to calculate the significance of each word using the concept of entropy, which quantifies the information carried by words. This enables us to assign appropriate weights to words, providing a more nuanced and accurate evaluation of how well the meaning has been preserved in the SR process.

Based on entropy, common words (high frequency, e.g., stop words) are considered to convey less information compared to less common words (low frequency, e.g., names of places and people).

The MScore function taking two text inputs, a complex text ($\mathcal{T}_C$) and a simplified text ($\mathcal{T}_S$), and returning a single score ($\mathcal{M}_{score}$) that signifies how faithfully $\mathcal{T}_S$ retains the original meaning of $\mathcal{T}_C$. Table 5 outlines the steps to calculate the $\mathcal{M}_{score}$ value, and here's a detailed breakdown of how the MScore Function operates:

It starts by tokenizing the $\mathcal{T}_C$ and storing the resulting tokens in the set $\mathcal{C}$ *(line 2)*. Similarly, it tokenizes the $\mathcal{T}_S$ and stores the tokens in the set $\mathcal{S}$. *(line 3)*. Then, It calculates the intersection set $\mathcal{I}$ by finding the common tokens between the set $\mathcal{C}$ and the set $\mathcal{S}$. These tokens in set $\mathcal{I}$ represent the words or phrases that are shared between both texts *(line 4)*. It also calculates the union set $\mathcal{U}$ by combining all the tokens from both sets $\mathcal{C}$ and $\mathcal{S}$. Set $\mathcal{U}$ represents all unique tokens from both texts *(line 5)*.

The $\mathcal{M}_{score}$ is calculated by dividing the sum of the significance weights of shared tokens by the sum of the significance weights for all unique tokens *(line 5)*. Here's how this is computed:

The $\mathcal{M}_{score}$ is calculated as a ratio of two sums. The numerator of the ratio is the sum of the reciprocals of $((1 + z(t))$ for all tokens *t* that are present in both $\mathcal{T}_C$ and $\mathcal{T}_S$ (in set $\mathcal{I}$). This value accounts for the 'Zipf-value'[3] for each common token[4][5]. The denominator of the ratio is the sum of the reciprocals of $((1 + z(t))$ for all tokens *t* that appear in either the $\mathcal{T}_C$ or the $\mathcal{T}_S$ (in set $\mathcal{U}$). This is calculated similarly to the numerator, but it includes all unique tokens from both $\mathcal{T}_C$ and $\mathcal{T}_S$.

Finally, the MScore Function returns the calculated $\mathcal{M}_{score}$ as its result *(line 7)*. The $\mathcal{M}_{score}$ is a single value, and its magnitude indicates the degree of M criterion in the range of 0 to 1. A higher $\mathcal{M}_{score}$ suggests a better preservation of meaning in the simplified text, while a lower score may indicate a reduction in meaning.

### 3.3 GScore Function

The evaluation of the G criterion for SR poses a formidable challenge, often considered one of intricate and complex tasks in NLP. This challenge stems from the intricate and interconnected nature of natural languages, which offer an endless array of possibilities and choices when it comes to sentence structures, word selections, and grammatical rules. Ensuring grammaticality in simplified text ($\mathcal{T}_S$) necessitates a profound understanding of complex linguistic interactions and the ability to adapt them into more accessible language.

---

[3] based on the SUBTLEX-US frequency list
[4] Essentially, it emphasizes the importance of less common words or tokens, as they tend to have lower values of $z(t)$
[5] For words that are not found in the SUBTLEX-US frequency list, the value of z(t) is set to 0.

To address the inherent difficulty in evaluating the G criterion, we adopted an approach that leverages the grammatical structure of the complex text ($\mathcal{T}_C$) as a reference point to estimate the G criterion of $\mathcal{T}_S$. This approach assumes that $\mathcal{T}_C$ is grammatically correct and uses it as a benchmark for evaluation. It involves breaking down $\mathcal{T}_S$ into n-gram set and measuring the precision of matching between this set and the corresponding n-gram set from $\mathcal{T}_C$. This evaluation process aims to quantify the extent to which $\mathcal{T}_S$ retains the grammatical structure and content of $\mathcal{T}_C$, providing a contextually informed assessment of the G criterion.

The longer n-gram matches yield more accurate estimations of grammaticality (Papineni and Roukos, 2002). However, in tasks like SR, which involves lexical simplification that primarily focus on substituting intricate vocabulary with simpler alternatives, directly identifying lengthy n-gram matches between $\mathcal{T}_S$ and $\mathcal{T}_C$ becomes more intricate. Consequently, addressing this challenge led us to devise a novel algorithm aimed at quantifying semi-matches within extended n-grams. This algorithm considers both full matches, where an entire n-gram from $\mathcal{T}_S$ aligns with a corresponding n-gram in $\mathcal{T}_C$, and partial matches, where (n-1) tokens in an n-gram from $\mathcal{T}_S$ correspond with (n-1) tokens in an n-gram from $\mathcal{T}_C$.

The objective of the **SemiMatch** function is to assess the presence of sequences of n tokens (n-grams) from a candidate text in a reference text. In Table 6, we outline the steps of the SemiMatch function, and in the following explanation, we dive deeper into these steps.

The SemiMatch function takes three inputs: *S*, representing the candidate text; *C*, representing the reference text; and *n*, which denotes the length of sequences of tokens *(line 1)*. It initializes two sets, $Cgram^n$ to store n-grams of the reference text *C* and $Sgram^n$ to store n-grams of the candidate text S *(lines 2 and 3)*.

The function calculates $M_{semi}$, representing the precision of n-grams from *S* that semi-match *C*, by using the $\partial$ function presented in Equation 2.

$$\partial(A^n, B^n) = \begin{cases} 0, & |A^n \cap^{\cdot} B^n| < n-1 \\ 1, & |A^n \cap^{\cdot} B^n| = n \\ \frac{n-2}{n}, & |A^n \cap^{\cdot} B^n| = n-1 \end{cases} \quad (2)$$

In this equation, $A^n$ and $B^n$ are n-grams (tuples), each consisting of n elements, typically words or phrases. $\cap^{\cdot}$ represents the intersection of two tuples, considering the order of elements.

The $\partial$ function calculates the intersection score based on the following conditions:

1. If the intersection of $A^n$ and $B^n$ contains fewer than $n-1$ common elements, the function returns 0, indicating a low degree of matching.

2. If the intersection contains exactly *n* common elements, the function returns 1, indicating a perfect match between $A^n$ and $B^n$.

3. If the intersection contains $n-1$ common elements, the function returns $\frac{n-2}{n}$, indicating a partial match.

The SemiMatch function calculates $M_{semi}$ by summing all the $\partial(cg_i^n, sg_j^n)$ values by iterating through all n-grams in $Cgram^n$ and $Sgram^n$ within nested loops, and then divides the sum by the total number of n-grams in the set $Sgram^n$, denoted as $|Sgram^n|$ *(line 4)*.

It's important to emphasize that we divide the sum by the total number of n-grams in the $Sgram^n$ set only, without considering the elements in the $Cgram^n$ set. This approach is taken because we are not estimating the overall similarity between *S* and *C*. Instead, our goal is to assess the extent to which sequences of words from *S* have a significant match with *C*. This nuance in the calculation allows us to focus on the precision of the n-gram semi-matching specifically from the candidate text to the reference text.

| Algorithm 6: SemiMatch Function | |
|---|---|
| 1: | **Function SemiMatch** ($C, S, n$) |
| 2: | $Cgram^n = \{cg^n | cg^n \in ngram(C)\}$ |
| 3: | $Sgram^n = \{sg^n | sg^n \in ngram(S)\}$ |
| 4: | $M_{semi} = \frac{\sum_{i=1} \sum_{j=1} \partial(cg_i^n, sg_j^n)}{|Sgram^n|}$ |
| 7: | **Return** $M_{semi}$ |

Table 6: shows the steps that the SemiMatch function performs to calculate $M_{semi}$

The GScore function operates with two text inputs: the original complex text ($\mathcal{T}_C$) and the simplified text ($\mathcal{T}_S$). It provides an estimation for the G criterion within a range of 0 to 1. To perform this evaluation, the GScore function leverages the *SemiMatch* function to measure how well the grammatical structure of $\mathcal{T}_S$ aligns with $\mathcal{T}_C$ as a reference, as outlined in Table 7.

The resulting GScore, denoted as $G_{score}$, is a numerical value that serves as an indicator of the grammatical correctness of $\mathcal{T}_S$. A higher $G_{score}$ typically suggests that $\mathcal{T}_S$ is more likely to be grammatically sound, while a lower score may indicate the presence of more grammatical errors or issues in $\mathcal{T}_S$.

The GScore function initializes a set $\mathbb{S}$ to store individual sentences obtained from $\mathcal{T}_S$. *(line 2)*. It calculates a set $\mathbb{S}M$ that contains average semi-match ($M_{semi}$) scores for each sentence $\mathbb{s}$ in $\mathbb{S}$ based on n-grams with *n* ranging from 4 to 7 *(line 3)*. The calculation for each sentence $\mathbb{s}$ is performed as follows:

- It iterates through n-grams with *n* ranging from 4 to 7.
- For each *n*, it calls the *SemiMatch* function with arguments $\mathcal{T}_S$, $\mathbb{s}$, and *n* to obtain a $M_{semi}$ score.
- The $M_{semi}$ scores for all n-grams are summed and divided by 4 to compute the average $M_{semi}$ score for the current sentence $\mathbb{s}$.
- This average $M_{semi}$ score is added to the $\mathbb{S}M$ set.

| Algorithm 7: GScore Function |  |
|---|---|
| 1: | **Function** $GScore$ ($\mathcal{T}_C, \mathcal{T}_S$) |
| 2: | $\mathbb{S} = \{\mathbb{s} \mid \mathbb{s} \in ToSentences(\mathcal{T}_S)\}$ |
| 3: | $\mathbb{S}M = \left\{\frac{1}{4}\sum_{n=4}^{7} \textbf{SemiMatch}(\mathcal{T}_C, \mathbb{s}, n) \mid \mathbb{s} \in \mathbb{S}\right\}$ |
| 4: | $G_{score} = \min_{sm \in \mathbb{S}M} \{sm \mid sm > 0\}$ |
| 5: | **Return** $G_{score}$ |

Table 7: shows the steps that the GScore function performs to calculate $G_{score}$

The GScore function calculates the $G_{score}$ by finding the minimum value from the $\mathbb{S}M$ set but only considering values greater than zero[6] *(line 5)*.

It's important to highlight that, based on our analysis of human evaluations, we found that calculating the $G_{score}$ by selecting the minimum value from the $\mathbb{S}M$ set correlates more strongly with human judgments of the G criterion than using the arithmetic mean of the $\mathbb{S}M$ set. This observation can be attributed to two key reasons:

Firstly, if a $\mathcal{T}_S$ contains a set of sentences that are all grammatically sound except for one ungrammatical sentence, the entire $\mathcal{T}_S$ should be considered to have failed the G criterion because it becomes practically unusable with one ungrammatical sentence. As a result, the $G_{score}$ for a $\mathcal{T}_S$ is based on the minimum score among all the sentences' scores.

Secondly, as the length of a sentence increases, the resulting $M_{semi}$ score tends to increase as well due to the increased likelihood of finding matches with $\mathcal{T}_C$. Consequently, the scores derived from shorter sentences are more indicative of the grammaticality of the $\mathcal{T}_S$.

## 4 Experimental Setup

In this experiment, we assess the accuracy of the CEScore model against conventional metrics currently employed for evaluating SR models. This comparative analysis is helpful in evaluating the viability of the CEScore model as an automated metric for the evaluation of SR models. We examine three widely used reference-based automatic metrics: BLEU, SARI, and BERTscore, as well as two reference-less automatic metrics designed for structural simplification: SAMSA and its variant, SAMSA<sub>abl</sub>.

We compare the scores from the metrics under consideration to the human evaluation scores from a sizable benchmark provided by Sulem et al. (2018 b)[7]. This benchmark encompasses human judgments regarding the performance of 26 models in simplifying the first 70 sentences of the TurkCorpus test set (Xu et al., 2016). These models cover a wide range of simplification transformations, covering both lexical and structural simplification. For lexical simplification, the models include:

1. Four versions of NTS, a state-of-the-art neural-based TS system (Nisioi et al., 2017) (NTS-h1, NTS-h4, NTS-h1 (w2v), and NTS-h4 (w2v))[8].
2. Two versions of the Moses TS system (Moses and Moses$_{LM}$) (Koehn et al., 2017).
3. SBMT-SARI a syntax-based machine translation system tuned against SARI (Xu et al., 2016).

---

[6] We excluded scores that equal zero because they often result from sentences that lack words, typically due to a punctuation error. In such cases, the function *ToSentences* (*in line 2*) may incorrectly split $\mathcal{T}_S$, leading to these zero scores.

[7] https://github.com/eliorsulem/simplification-acl2018

For structural simplification, the modes include:

1. HYBRID system that utilizes semantic structures for sentence splitting and deletion (Narayan and Gardent, 2014).
2. Two versions of DSS, a model for sentence splitting based on a semantic parser (DSS and DSS$^m$)[9] (Sulem et al., 2018 c).
3. Eight versions of SENTS, a system that combines structural and lexical simplification with the NTS model implemented on top of DSS (SENTS-h1, SENTS-h4, SENTS-h1 (w2v), SENTS-h4 (w2v), SENTS-h1$^m$, SENTS-h4$^m$, SENTS-h1$^m$ (w2v), and SENTS-h4$^m$ (w2v))[10] (Sulem et al., 2018 c).
4. Eight versions of SEMoses, where Moses, a phrase-based machine translation system, is implemented on top of DSS (SEMoses, SEMoses$^m$, SEMoses$_{LM}$, SEMoses$_{LM}^m$, SETrain1-Moses, SETrain1-Moses$_{LM}$, SETrain2-Moses, SETrain2-Moses$_{LM}$) (Sulem et al., 2018 c).

In total, our study involved the human evaluation of 1820 sentences from 26 TS models. Each of these sentences evaluated by three native English annotators based on four criteria: Meaning preservation (M), Grammaticality (G), Simplicity (S), and Structural Simplicity (StS) [11]. The criteria G and M were rated on a scale ranging from 1 to 5, while S and StS were evaluated on a scale spanning from -2 to 2 [15].

For the reference-based automatic metrics, we utilized the HSplit corpus (Sulem et al., 2018 a)[12], which comprises four different versions of human-generated simplifications tuned for SR task.

We conducted a comprehensive comparative analysis between CEScore and the considered automatic metrics across four criteria: S[13], G, M, and overall quality. For each of these criteria, we compared the scores obtained from the automatic metrics with the human judgment scores corresponding to that specific criterion.

In the case of the overall quality criterion, we determined the overall quality using the approach introduced by AlAjlouni et al. (2023). This approach utilizes their formula known as "Standard A," which was found to be more accurate than the traditional arithmetic mean in combining the G, M, and S criteria to reflect the overall quality of SR task. This choice was made based on their observations of correlations in human evaluations, particularly between G and S, as well as between G and M. To provide a comprehensive view, we also reported the overall quality based on the arithmetic mean.

We conduct this comparative analysis on two levels. First, at the sentence-level, we aim to assess the accuracy of CEScore in evaluating the quality of simplification for individual sentences. We compare the scores generated by CEScore and other automatic metrics with the corresponding human evaluations for each sentence, considering all models. In essence, for each evaluation criterion, we compare the outcomes of the automatic metrics for 1,820 sentences with their corresponding human assessments.

Second, at the model (system) level, we aim to gauge the effectiveness of the CEScore in evaluating the models as a whole. This level of analysis focuses on the average ratings for each model. For each criterion, we compare the corpus-scores provided by CEScore and other automatic metrics for 26 models with the average human scores for each model in each criterion.

---

[8] Either the system relies on its default sitting or initialized with Word2Vec. In both cases, the top-ranked and fourth-ranked hypotheses are taken.
[9] DSS$^m$ is a semi-automatic version of DSS, which relies on manual annotation instead of the parser.
[10] X$^m$ refers to the semi-automatic version of the system X, where DSS$^m$ is used instead of DSS.

[11] In the context of structural simplicity (StS), annotators are explicitly directed to disregard lexical simplicity, which involves the replacement of complex terms with simpler, more common ones.
[12] https://github.com/eliorsulem/HSplit-corpus
[13] the structural simplicity (StS) less detailed and focused compared to the broader criterion of simplicity.

# 5 Results

We conducted a comprehensive comparison of $S_{score}$ and the considered automatic metrics against human judgments for the S criterion. As shown in table 8, The results showed that $S_{score}$ displayed strong correlations with human judgment scores at both sentence and model levels. At the sentence level, the correlation coefficients were 0.64 and 0.56 for Pearson and Spearman, respectively. Remarkably, at the model level, $S_{score}$ exhibited exceptionally strong correlations, with Pearson's and Spearman's coefficients scoring at 0.86 and 0.83, respectively. This strong positive correlation is a noteworthy finding, especially when compared to other automatic metrics, both reference-based and reference-less.

Surprisingly, automatic metrics designed to be positively correlated with simplicity, such as SARI, SAMSA, and SAMSA$_{abl}$, displayed an unexpected negative correlation with human judgment under the S criterion. This surprising result may be attributed to the fact that SARI is primarily designed to estimate lexical TS only, while SAMSA and SAMSA$_{abl}$ are tailored for assessing structural TS only. In contrast, the models in our experiment cover both lexical and structural simplification, which could explain the discrepancies in correlation. Figure 1 displays scatter plots that depict the relationship at the sentence-level between human judgment scores for S criterion and the corresponding scores from automatic metrics. Each data point in these plots corresponds to an individual sentence, and regression lines are included for reference.

In the case of G criterion, $G_{score}$ outperformed the other atomic metrics with notable margin at both sentence and model levels, as shown in table 9. At model-level, $G_{score}$ demonstrated very strong correlations, with 0.89 and 0.85 for Pearson's and Spearman's ρ, respectively. At sentence level, the correlation between $G_{score}$ and human judgment was moderate, with Pearson and Spearman ρ of 0.55 and 0.53, respectively. However, Evaluating G is undoubtedly a complex process influenced by various factors, making the evaluation challenging and increasing the likelihood of incorrect assessments. Figure 2 illustrates scatter plots that reveal the relationship at the model-level between human judgment scores for the G criterion and the corresponding scores from automatic metrics. Each data point in these plots corresponds to an individual model.

For the M criterion, BLEU and BERTscore exhibited very strong correlations with human judgments, outperforming other automatic metrics. At the sentence level, BERTscore achieved the highest correlation with human judgment, with

| | Metric | Sentence-Level | | Model-Level | |
|---|---|---|---|---|---|
| | | Pearson | Spearman | Pearson | Spearman |
| Reference-based | BLEU | -0.37 | -0.26 | -0.69 | -0.69 |
| | SARI | -0.29 | -0.27 | -0.80 | -0.78 |
| | BERTscore | -0.44 | -0.33 | -0.78 | -0.73 |
| Reference-less | SAMSA | -0.27 | -0.27 | -0.72 | -0.68 |
| | SAMSA$_{abl}$ | -0.32 | -0.33 | -0.51(0.01) | -0.47(0.01) |
| | $S_{score}$ | **0.64** | **0.56** | **0.86** | **0.83** |

Table 8: The Spearman's and Pearson's coefficients, and their associated p-values (if greater than 0.005) for the correlations between the automatic metrics and human judgment under the **S criterion**, at both sentence and model levels. The highest score in each column is denoted in bold.

| | Metric | Sentence-Level | | Model-Level | |
|---|---|---|---|---|---|
| | | Pearson | Spearman | Pearson | Spearman |
| Reference-based | BLEU | 0.41 | 0.47 | 0.64 | 0.59 |
| | SARI | 0.01 | -0.08 | 0.09 | -0.15(0.45) |
| | BERTscore | 0.39 | 0.44 | 0.52 | 0.55 |
| Reference-less | SAMSA | 0.20 | 0.20 | -0.04(0.86) | -0.01(0.96) |
| | SAMSA$_{abl}$ | 0.34 | 0.33 | 0.58 | 0.50(0.01) |
| | $G_{score}$ | **0.55** | **0.53** | **0.89** | **0.85** |

Table 9: The Spearman's and Pearson's coefficients, and their associated p-values (if greater than 0.005) for the correlations between the automatic metrics and human judgment under the **G criterion**, at both sentence and model levels. The highest score in each column is denoted in bold.

| | Metric | Sentence-Level | | Model-Level | |
|---|---|---|---|---|---|
| | | Pearson | Spearman | Pearson | Spearman |
| Reference-based | BLEU | 0.77 | 0.75 | **0.98** | **0.95** |
| | SARI | 0.36 | 0.24 | 0.79 | 0.63 |
| | BERTscore | **0.84** | **0.83** | 0.97 | 0.94 |
| Reference-less | SAMSA | 0.45 | 0.49 | 0.72 | 0.77 |
| | SAMSA$_{abl}$ | 0.58 | 0.64 | 0.80 | 0.79 |
| | $M_{score}$ | 0.79 | 0.75 | 0.94 | 0.94 |

Table 10: The Spearman's and Pearson's coefficients, and their associated p-values (if greater than 0.005) for the correlations between the automatic metrics and human judgment under the **M criterion**, at both sentence and model levels. The highest score in each column is denoted in bold.

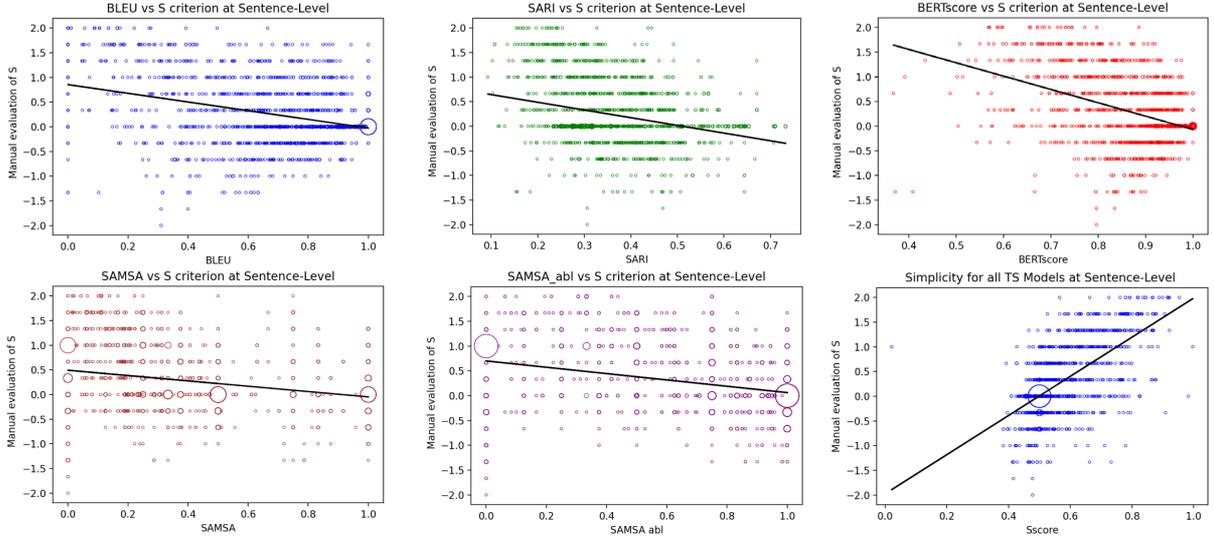

Figure 1: Scatter plots and regression lines at the sentence-level, depicting the relationship between the human scores assigned for S criterion (Y-axes) and the corresponding automatic model scores (X-axes). Each data point on the graph represents a sentence. (The size of each data point corresponds to the number of repetitions)

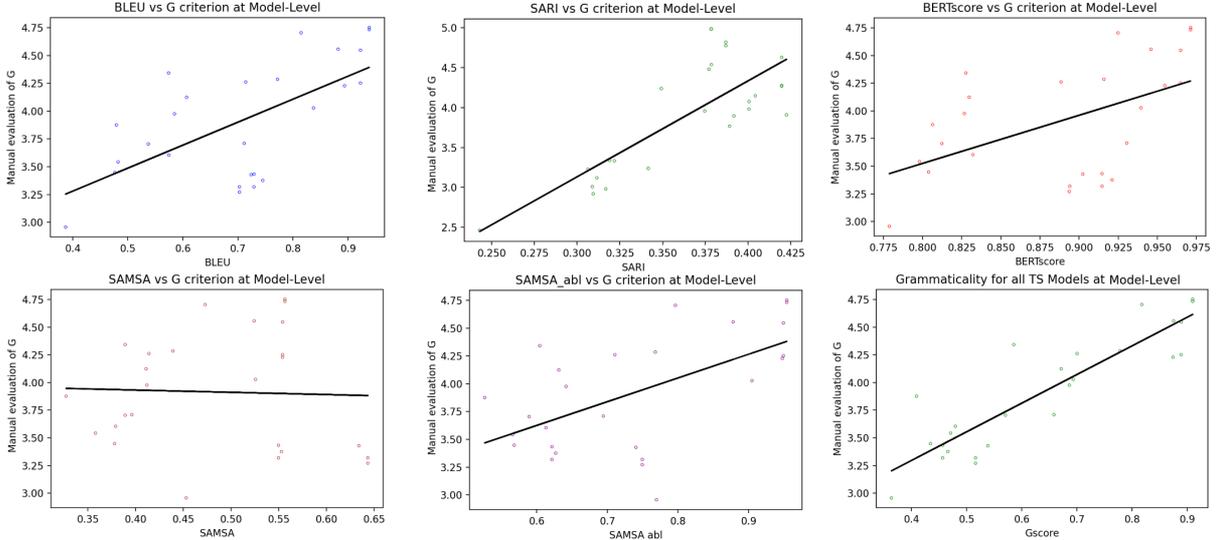

Figure 2: Scatter plots and regression lines at the model-level, depicting the relationship between the human scores for G assigned the Y-axes and the corresponding automatic model scores on the X-axes. Each data point on the graph represents a model.

Pearson's and Spearman's correlation coefficients of 0.84 and 0.83, respectively. $M_{score}$ came second, with Pearson's and Spearman's correlation coefficients of 0.79 and 0.75, respectively.

At the model level, BLEU demonstrated remarkable performance, with Pearson's and Spearman's correlation coefficients of 0.98 and 0.95, respectively. $M_{score}$ also showed a very strong correlation with human judgment, scoring 0.94 for both Pearson and Spearman correlation coefficients. This is a notable achievement, considering that $M_{score}$ is a reference-less metric, while both BLEU and BERTscore rely on the estimation based on well-prepared references. In comparison to SAMSA and SAMSA$_{abl}$, the reference-less metrics in our experiment, $M_{score}$ exhibited impressive performance.

In terms of overall quality, we conducted a comprehensive comparison of the automatic metrics against overall quality, which we calculated using two methods: first, by following AlAjlouni's methodology, utilizing 'Standard A' ($F_A$); second, by employing the more conventional approach of relying on the arithmetic mean of M, G, and S scores ($F_{avg}$).

|  | Metric | Sentence-Level | | | | Model-Level | | | |
|---|---|---|---|---|---|---|---|---|---|
|  |  | Pearson | | Spearman | | Pearson | | Spearman | |
|  |  | $F_A$ | $F_{avg}$ | $F_A$ | $F_{avg}$ | $F_A$ | $F_{avg}$ | $F_A$ | $F_{avg}$ |
| Reference-based | BLEU | 0.61 | 0.55 | 0.56 | 0.63 | 0.88 | 0.85 | 0.85 | 0.83 |
| | SARI | 0.23 | 0.13 | 0.17 | 0.04 | 0.52 | 0.38 (0.05) | 0.31 (0.12) | 0.21 (0.29) |
| | BERTscore | **0.62** | 0.56 | **0.56** | **0.64** | 0.81 | 0.75 | 0.82 | 0.79 |
| Reference-less | SAMSA | 0.30 | 0.29 | 0.32 | 0.28 | 0.34 (0.09) | 0.29 (0.15) | 0.43 (0.03) | 0.48 (0.01) |
| | SAMSA$_{abl}$ | 0.43 | 0.42 | 0.46 | 0.37 | 0.72 | 0.73 | 0.78 | 0.81 |
| | $CE_{score}$ | 0.60 | **0.57** | 0.53 | 0.59 | **0.95** | **0.94** | **0.98** | **0.95** |

Table 11: The Spearman's and Pearson's coefficients, along with their associated p-values (if greater than 0.005), represent the correlations between the automatic metrics and the overall quality based on human judgments, examined at both the sentence and model levels. The overall quality was calculated using two methods: AlAjlouni's 'Standard A' ($F_A$) or the arithmetic mean ($F_{avg}$). The highest score in each column is denoted in bold.

At the sentence-level, BLEU, BERTscore, SAMSAabl, and $CE_{score}$ exhibited a moderate correlation with overall quality, whether by $F_A$ or $F_{avg}$, as shown in Table 11. BERTscore took the lead, followed closely by BLEU, and $CE_{score}$ followed in third place. Notably, the margins between their performance were quite small, and $CE_{score}$ even outperformed the others in some cases (see Table 11).

At the model-level, all metrics (except SARI and SAMSA) showed very strong correlations with overall quality, whether calculated using $F_A$ or $F_{avg}$. Notably, $CE_{score}$ displayed an impressive correlation with $F_A$, achieving coefficient values of 0.95 and 0.98 for Pearson and Spearman, respectively. The correlation with $F_{avg}$ was also substantial, with values of 0.94 and 0.95 for Pearson and Spearman, respectively. These results underscore the effectiveness of using $CE_{score}$ as a robust metric for evaluating SR at the model-level.

Figure 3 presents scatter plots illustrating the relationship at the sentence-level between the values of automatic metrics and the corresponding overall quality scores based on $F_{avg}$. Each data point on these plots corresponds to an individual sentence. Additionally, Figure 4 showcases scatter plots at the model-level between automatic metrics and $F_A$, where each point represents an individual model. The regression lines are included in these graphs for reference.

## 6 Discussion

The metrics of the CEscore model, including $S_{score}$, $G_{score}$, $M_{score}$, and $CE_{score}$, exhibited strong correlations with human judgments across all the aspects under consideration, especially at the model level. This suggests the model's effectiveness in evaluating SR models, surpassing other automatic metrics. This is because each metric in CEscore model is designed to measure specific criteria in a way similar to human evaluation.

The experiment highlighted the success of $S_{score}$ in estimating the S criterion and $M_{score}$ in estimating the M criterion at the sentence level. Both of these metrics demonstrated strong correlations with human judgments, outperforming all other automated metrics.

The significant difference between the results of SAMSA and SAMSA$_{abl}$ highlights the limitations of the semantic parser in accurately converting sentences into scenes, which had a considerable impact on their evaluation accuracy.

SARI performed poorly in the experiment, as it did not show a strong correlation with any of the criteria under consideration. This may be attributed to SARI being designed for measuring only a specific type of simplification, which is lexical simplification. Its performance is significantly reduced when the simplified text contains any form of structural simplification, making it unsuitable for evaluating SR task.

Both BLEU and BERTscore metrics demonstrated excellent performance in evaluating meaning preservation and overall quality at the model level. This emphasizes their suitability for evaluating the overall quality of SR models, provided there are sufficient and diverse references available.

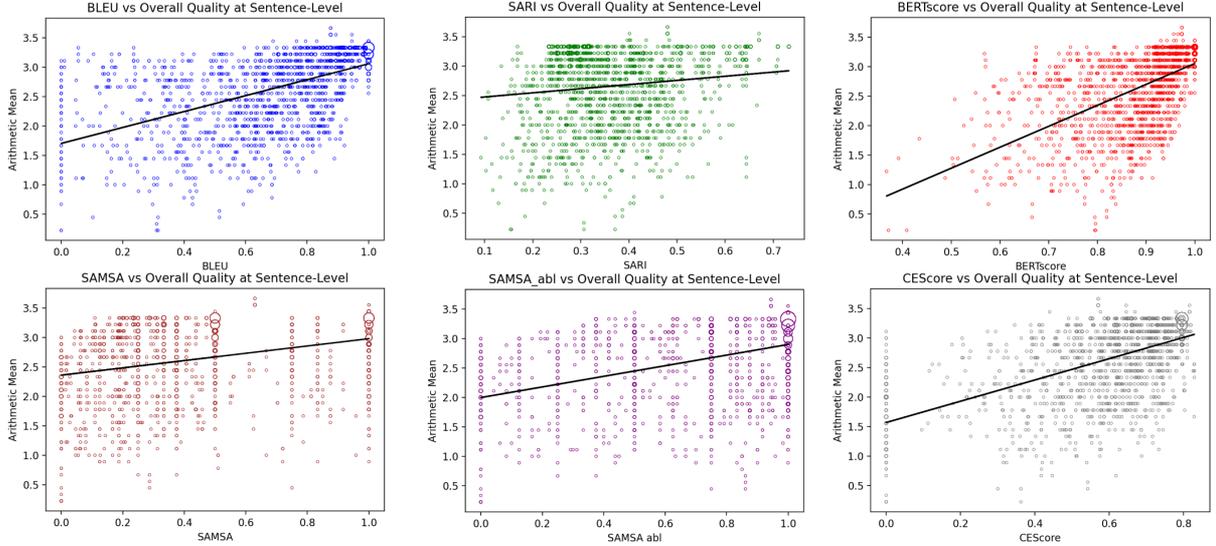

Figure 3: This figure includes scatter plots and associated regression lines at the sentence-level, illustrating the relationship between overall quality calculated by $F_{avg}$ (Y-axes) and the corresponding scores of the automatic models (X-axes). Each data point on the graph represents an individual sentence, with the size of each data point corresponding to the number of repetitions.

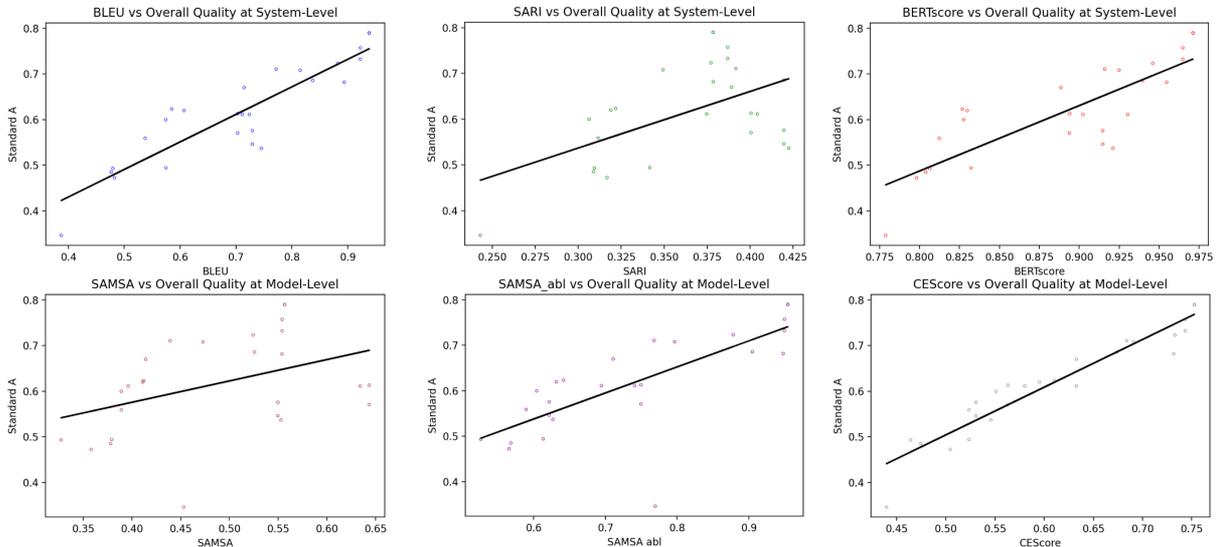

Figure 4: This figure includes scatter plots with regression lines as references at the model-level, illustrating the relationship between overall quality as calculated by $F_A$ (Y-axes) and the corresponding scores of the automatic models (X-axes). Each data point on the graph represents an individual model.

## 7 Conclusion

In conclusion, this research addresses the critical need for precise and efficient evaluation metrics in the field of SR task. Traditional evaluation metrics, originally designed for machine translation tasks, fall short in capturing the unique goals and challenges of SR. As a response to these limitations, we introduce CEScore, a Confidence Estimation Score model that directly evaluates the quality of simplified text in terms of simplicity (S), meaning preservation (M), and grammaticality (G). CEScore's comprehensive approach aligns with human evaluation and eliminates the reliance on reference texts, making it a valuable and contextually relevant tool for SR evaluation.

Our research contributes to the field in several key aspects. We have introduced statistical functions and innovative formulas for evaluating simplicity, including Sentence Length Score (SLS), Average Sentence Familiarity (ASF), and Text Simplicity Score (TSS). These tools provide valuable insights into the simplicity of SR models' output. Additionally, CEScore itself, which includes $S_{score}$, $M_{score}$, $G_{score}$, and $CE_{score}$, demonstrates strong correlations with human judgments. Particularly at the model level, it achieves remarkable coefficient values, making it a reliable and effective evaluation metric for SR models.

Moving forward, future work will focus on enhancing the accuracy of $G_{score}$ and $M_{score}$ by incorporating contextual embeddings. This step will address their sensitivity to lexical simplification, further improving their ability to evaluate SR models accurately. These developments will contribute to the ongoing advancement of the field of SR and evaluation, ultimately making digital content more accessible and comprehensible to a broader audience.